%% file: IS17.tex
\documentclass[a4paper]{article}
\usepackage{INTERSPEECH_v2}

\usepackage{times}
\usepackage{subfigure}
\usepackage{graphicx}
\usepackage{latexsym}
\usepackage{comment}
\usepackage{calrsfs}
\usepackage{changepage}
\usepackage{lipsum}
\usepackage{amsmath}
\usepackage{multirow}
\usepackage{url}
\usepackage{adjustbox}
\usepackage{placeins}
\DeclareMathAlphabet{\pazocal}{OMS}{zplm}{m}{n}

\title{Joint Learning of Correlated Sequence Labeling Tasks Using Bidirectional Recurrent Neural Networks}
\name{Vardaan Pahuja${}^{1*}$, Anirban Laha${}^{1*}$\thanks{*Equal contribution by the first two authors.}, Shachar Mirkin${}^2$, Vikas Raykar${}^1$, Lili Kotlerman${}^2$, Guy Lev${}^2$}
\address{
  ${}^{1} $IBM Research - Bangalore, India\\
  ${}^{2} $IBM Research - Haifa, Israel}
\email{vapahuja@in.ibm.com, anirlaha@in.ibm.com, shacharm@il.ibm.com, viraykar@in.ibm.com, lili.kotlerman@il.ibm.com, guylev@il.ibm.com}

\begin{document}

\maketitle

\input{abstract}

\input{intro}

\input{seq}

\input{joint}

\vspace*{-2mm}
\input{exp}

\input{tables}
\input{results}

\input{related}

\vspace*{-4mm}

\input{conclusion}








\bibliographystyle{IEEEtran}
\bibliography{IS17}

\end{document}

%% file: abstract.tex
\begin{abstract}
The stream of words produced by Automatic Speech Recognition (ASR) systems is typically devoid of punctuations and formatting. Most natural language processing applications expect segmented and well-formatted texts as input, which is not available in ASR output. This paper proposes a novel technique of jointly modeling multiple correlated tasks such as punctuation and capitalization using bidirectional recurrent neural networks, which leads to improved performance for each of these tasks. This method could be extended for joint modeling of any other correlated sequence labeling tasks.
\end{abstract}

%% file: intro.tex
\section{Introduction}
\emph{Sequence labeling} involves the assignment of a categorical label to each element of a sequence of tokens. Some common examples include punctuation prediction for automatic speech recognition (ASR) transcripts, capitalization recovery (i.e. restoring the case of the lowercased words, a.k.a. \textit{truecasing}), part-of-speech tagging (POS), and named entity recognition (NER). In this work we address the task of \emph{multiple sequence labeling}, where the goal is to assign \emph{multiple categorical labels} to each element of the sequence, such as predicting both the punctuation and capitalization for a given ASR speech transcript. We specifically address the scenario in which the multiple sequence labeling tasks are correlated. Consider the following two examples:
\begin{enumerate}
    \item \small \textit{\dots{}and it hasn't been refined enough yet\underline{\textbf{. It}} needs to be worked on until it can speak fluently}
    \item \textit{This young doctor\underline{\textbf{, Tom Ferguson,}} was the medical editor of the Whole Earth Catalog.}
\end{enumerate}
The first example shows the occurrence of capitalization preceded by a period. In the second example, the two commas surround capitalized proper nouns. Such co-occurrences illustrate the fact that punctuation and capitalization are two correlated tasks that could benefit from each other. We refer to these kinds of sequence labeling tasks as \emph{correlated multiple sequence labeling} and propose a novel approach using a bidirectional recurrent neural network (BiRNN) \cite{birnn97}, that is trained jointly for prediction across such tasks.

Speeches are often transcribed by ASR systems that convert the audio signals into a stream of words. Apart from often having a high word error rate, this stream is also devoid of the standard textual structure present in written texts. These structural aspects \cite{liu2006enriching} include punctuation, capitalization, and numeric data formatting, such as for digits, dates, and phone numbers.
%
%
Recovering the structure from raw word transcripts is essential for two main reasons. First, the structure enhances the readability and understanding of the transcripts \cite{shugrina10,tilk2015lstm}. Second, its recovery enables subsequent text processing and makes it more accurate. Many works have shown the impact of the structure recovery for tasks such as summarization \cite{liu2008impact,mrozinski2006automatic}, part-of-speech (POS) tagging \cite{hillard2006impact,lita2003truecasing}, named entity recognition (NER) \cite{hillard2006impact}, machine translation \cite{paulik2008sentence,matusov2006automatic} and information extraction \cite{favre2008punctuating}, among others.

We attempt to recover two aspects of structure, punctuation and capitalization, by casting it as correlated multiple sequence labeling problem. 
%
%
%
Earlier work \cite{collobert2011natural} proposed the idea of training multiple sequence labeling tasks together, and showed a slight improvement for POS and NER when combined with task-specific feature engineering. However, they assumed the availability of sentence segmentation and capitalization as inputs. The solution we propose does not assume any feature engineering and is suitable for speech transcripts, which  do not come with punctuation or capitalization.

Multiple papers \cite{liu2006enriching,tilk2015lstm,kolar2012development,kim2001use,huang2002maximum,levy2012effect,baron2002automatic,eidelman10} showed the usefulness of pause duration and prosodic features for punctuation prediction as compared to using textual features alone. 
In this work, our goal is to boost the accuracy of punctuation prediction without taking into account additional inputs such as prosodic features; we accomplish this by training the capitalization task jointly with the punctuation task. 
To the best of our knowledge, this is the first RNN (BiRNN)-based framework for joint training of correlated sequence labeling tasks. Moreover, this framework is general enough to be applicable for jointly training other correlated sequence labeling tasks such as POS tagging and NER.

In a nutshell, our contributions are the following:

\begin{itemize}
    \item An RNN (BiRNN)-based joint learning framework for multiple correlated sequence labeling tasks, with no feature engineering.
    \item Improvement in punctuation prediction on speech transcripts by jointly training it with capitalization, without using any prosodic features. A similar improvement is also observed in capitalization.
    \item State-of-the-art performance on benchmark punctuation prediction dataset.
\end{itemize}

%% file: seq.tex
\section{Correlated Multiple Sequence Labeling}
\label{sec:seq}
Punctuation and capitalization are considered highly important for the structure recovery of ASR transcripts. There are various effective approaches to insert punctuation and specifically sentence boundaries into raw speech transcripts \cite{wang2012dynamic,xu2014deep}. In this work, we consider both punctuation and capitalization together, treating it as a \textit{correlated multiple sequence labeling} problem, as defined below:

\begin{adjustwidth}{2em}{0pt}
\textit{Given a sequence of words $W=(w_1,w_2,w_3,...,w_n)$ from a vocabulary $V$, the objective is to predict $K$ labels $\{l_i^1,l_i^2, ... ,l_i^K\}$ corresponding to each word $w_i$, one label for each of the $K$ sequence labeling tasks. This will produce $K$ correlated output sequences of the form $O^k=(l_1^k,l_2^k, ... ,l_n^k)$. Here, labels for different tasks come from different label spaces, as in $l_i^k \in L^k$.}

\end{adjustwidth}
Following the above definition, $K=1$ trivially implies a \textit{single sequence labeling problem}. In our setting, $K=2$ when we consider the punctuation and capitalization tasks together. Typically, three punctuation marks have received the most attention in existing literature due to their high frequency of occurrence: periods, commas, and question marks (Q-MARK below). Thus, $L^1=$\{COMMA, PERIOD, Q-MARK, NO-PUNCT\}, where there is a high class imbalance tilted towards the NO-PUNCT class. In our model formulation, the label $l_i^1$ corresponds to the punctuation occurring before the word $w_i$. As for capitalization, the label $l_i^2$ determines the surface form of word $w_i$, which can be any of the following: all-lowercase (e.g., `hello'), all-uppercase (e.g.,  `NASA'), mixed-case (e.g., `McGill'), sentence-case (only the first letter capitalized, e.g., `London') and single-letter-word-case (e.g., `I').

%% file: joint.tex
\begin{figure}
\raggedleft
\includegraphics[width=1.0\columnwidth]{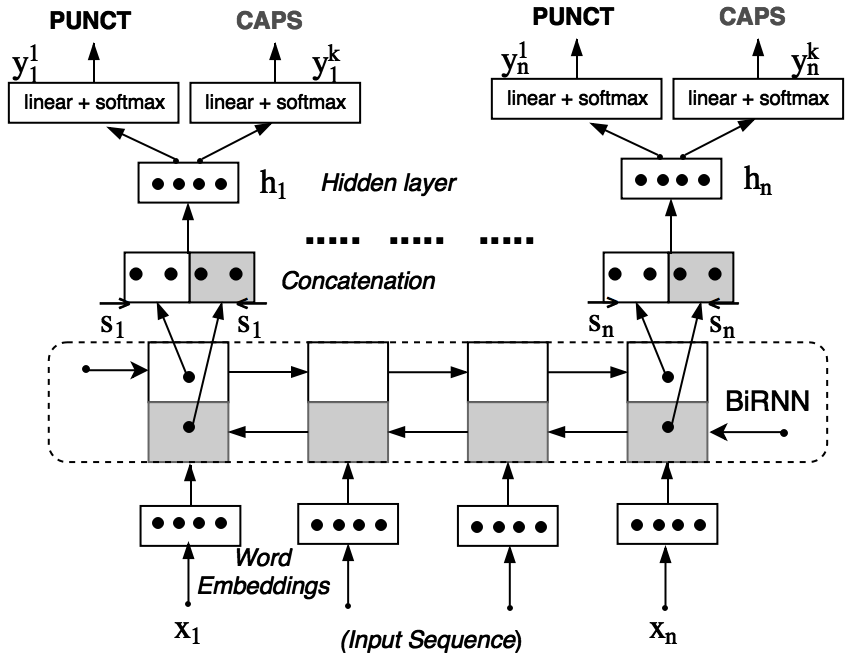}
\caption{Framework for Correlated Sequence labeling Tasks}\label{fig:models}
\end{figure}


Given a sequence of $n$ input vectors $\textbf{x}_1,...,\textbf{x}_n$ and an initial state vector $\textbf{s}_0$, an RNN generates a sequence of $n$ state vectors $\textbf{s}_1,...,\textbf{s}_n$ alongside a sequence of $n$ output vectors $\textbf{y}_1,...,\textbf{y}_n$; that is, $RNN(\textbf{s}_0,\textbf{x}_1,...,\textbf{x}_n)=\textbf{s}_1,...,\textbf{s}_n,\textbf{y}_1,...,\textbf{y}_n$. The input vectors $\textbf{x}_i$ are the latent embeddings (word2vec \cite{word2vec13}) of the words $w_i$ in the sequence, and $\textbf{s}_i$ represents the state of the RNN after observing the inputs $\textbf{x}_1,...,\textbf{x}_i$. The output vector $\textbf{y}_i$ is a function of the corresponding state vector $\textbf{s}_i$ and is then used for prediction of output labels for the correlated tasks. An RNN is defined by the following update equations: $\textbf{s}_i  =  R(\textbf{x}_i,\textbf{s}_{i-1})$ and $\textbf{y}_i  =  O(\textbf{s}_{i})$.
Different instantiations of $R$ and $O$ will result in different network structures (Simple RNN, LSTM~\cite{lstm97}, GRU~\cite{cho14}, etc.). 

A bidirectional RNN consists of two parallel RNNs: one running forward and another running backward. These capture the context in both directions (since the words to the right have significant influence on a word label in addition to the words to its left). Essentially, the same sequence of input vectors $\textbf{x}_1,...,\textbf{x}_n$ is fed to both RNNs to produce the sequence of state vectors  $\overrightarrow{\textbf{s}_1},...,\overrightarrow{\textbf{s}_n}$ from the forward RNN and  $\overleftarrow{\textbf{s}_1},...,\overleftarrow{\textbf{s}_n}$ from the backward RNN. Here we extend the bidirectional RNN to model multiple correlated sequence labeling tasks together. For the $k$-th task being considered, the output sequence, denoted by $\textbf{y}^k_1,...,\textbf{y}^k_n$, can be derived from the sequence of state vectors $\textbf{s}_1,...,\textbf{s}_n$, where $\textbf{s}_i=[\overrightarrow{\textbf{s}_i},\overleftarrow{\textbf{s}_i}]$, through transformations, as defined below:
\begin{eqnarray}
  \textbf{h}_i = f(\textbf{s}_{i}) = \phi(\textbf{W}  \textbf{s}_{i} + \textbf{b}) \\
  \textbf{m}^k_i = g^k(\textbf{h}_i) = \textbf{W}^k \textbf{h}_{i} + \textbf{b}^k \\
  \textbf{y}^k_i = softmax(\textbf{m}^k_i) 
\end{eqnarray}
In the above formulation, the concatenated state vector $\textbf{s}_i$ is transformed linearly and passed through the function $\phi \in \{sigmoid,tanh,relu,linear\}$ to produce a hidden layer vector $\textbf{h}_{i}$. To produce outputs for the different correlated tasks in question, the vector $\textbf{h}_{i}$ is then passed through different branches of $linear+softmax$, one branch for each of the tasks. 
That is, for the $k$-th task, the output $\textbf{y}^k_i$, is produced from the $k$-th branch, which leads to the prediction of label $l^k_i$.
The set of trainable parameters are $\{\textbf{W},\textbf{b},\{\textbf{W}^k,\textbf{b}^k\}_{k=1}^{K}\}$ in addition to the parameters defining the forward and backward RNNs. Figure~\ref{fig:models} illustrates our model formulation. 

\textbf{Joint Training Loss Function:} The network formulated above is defined for multiple correlated tasks (say $K$ tasks) and is capable of producing $K$ sequences of outputs of the form $\textbf{y}^k_1,...,\textbf{y}^k_n$. While predicting the outputs for the different tasks, all the trainable parameters required until the computation of $\textbf{h}_i$ are shared across all tasks and are trained jointly based on the loss function defined over the outputs of all $K$ tasks. We compute the loss $\pazocal{L}^k$ for every task using the standard cross-entropy loss function. Then, based on predefined weights $q_k$ (over tasks), a weighted average of task-specific losses is taken to produce the final loss $\pazocal{L}$ to be optimized:
\vspace*{-3mm}
\begin{eqnarray}
\pazocal{L} = \sum_{k=1}^{K} q_k \pazocal{L}^k
\end{eqnarray}
\vspace*{-3mm}

This accumulated loss helps the network predict well across all tasks. If the tasks are correlated (as in our case), then each task should help the other tasks through the joint learning of shared parameters. These shared parameters help produce \textit{correlated representations} $\textbf{h}_i$, which can be used to generate predictions for all tasks. 

%% file: exp.tex
\newcommand{\specialcell}[2][c]{%
  \begin{tabular}[#1]{@{}c@{}}#2\end{tabular}}
  
\section{Experiments}
\label{sec:exp}

To corroborate the hypothesis that our jointly trained model helps improve performance over the individual tasks, we experimented on two different datasets, as described below. 
All our models\footnote{The code will be available at \url{https://goo.gl/3UGd4p}} are evaluated based on precision (P), recall (R) and F$_1$ score, for each punctuation class, and overall for all classes, as well as with Slot Error Rate (SER)\footnote{SER is the ratio of the total number of slot errors (substitutions, deletions, and insertions) in the predicted set of labels, to the total number of slots in the gold set of labels.} \cite{makhoul1999performance}. 
\vspace*{-2mm}
\subsection{Datasets}
\textbf{Intelligence Squared}: This dataset was obtained from the Intelligence Squared (IQ2 henceforth) debating television show, whose transcripts are publicly available.\footnote{\url{http://www.intelligencesquaredus.org/}} We used 45 debates, each containing talks by four speakers, from which we created a train-validation-test split in a ratio of 60:10:30.\footnote{Evaluated on reference transcripts only as ASR is not available.}

\textbf{IWSLT TED Talks}: We used the English transcripts of the English-to-French machine translation task in IWSLT 2012\footnote{\url{https://wit3.fbk.eu/mt.php?release=2012-03}} as our training data with the same train-validation splits as suggested in \cite{lrec16}. We report our test results on two datasets\footnote{Both Reference transcripts and ASR}: the first, used by \cite{ueffing13} (henceforth referred to as test-set-1), is the development data of the IWSLT 2011 ASR and SLT tasks\footnote{\url{http://iwslt2011.org/doku.php?id=06_evaluation}}; the second test set (henceforth referred to as test-set-2) consists of test-dataset-2 of the IWSLT 2011 ASR and SLT tasks, as used by \cite{lrec16} and T-BRNN \cite{tilk2016bidirectional}. 



\subsection{Experimental Setup}
\textbf{Data Preprocessing}: 
Each training sequence consists of a random number of tokens (40 to 70 in our experiments), with the constraint that it must begin with a new sentence. The unfinished sentence forms the beginning of next training sequence. This scheme of generating training sequences prevents the model from always learning to predict a period or a question mark at the end of every sequence. For the validation and test datasets, we used a single consolidated sequence comprising all the sentences, to simulate a real ASR stream. This is not done for the training dataset to avoid memory issues with extremely long sequences. 
To evaluate our model on ASR transcripts, we mapped the punctuations and capitalization from the reference transcripts to the ASR transcripts, based on Levenshtein alignment, as discussed in \cite{kolar2012development}. 
Since the mapping process is sensitive to ASR word errors, we adopted the approach in \cite{ueffing13}, and restricted the evaluation to only those punctuations for which the left and right context words have been recognized correctly by the ASR. Similarly, we restricted capitalization evaluation to the words matching in the reference. For punctuation, we used the standard four classes as mentioned in Section~\ref{sec:seq}, whereas for capitalization, sentence-case and mixed-case were merged as the latter occurs very rarely.

\textbf{Network Training and Tuning}: We trained our model architecture using standard backpropagation in TensorFlow \cite{abadi2016tensorflow}. 
In our experiments, we trained two kinds of models: joint model (or Corr-BiRNN), which was trained jointly on punctuation and capitalization tasks, and task-specific models (or Single-BiRNN), that were trained separately for each of the two tasks. 
We carried out extensive hyper-parameter tuning for both the joint model and the separate task-specific models, for the IQ2 and TED datasets. The tuned hyper-parameters included: the number of layers and the number of hidden units per layer in the BiRNN, RNN dropout rate, RNN output dropout rate, type of RNN (Simple RNN, LSTM or GRU), the number of units in the outer hidden layer, hidden layer activation function, task-specific loss weights, and batch size.
The best hyper-parameter setting for the joint model as well as the task-specific models was selected based on SER performance on the validation set for the task at hand.
We then evaluated the selected settings on the reference transcripts of the corresponding test sets and on ASR test set (available for TED only) for the respective tasks.
Note that the best hyper-parameter setting for a punctuation task-specific model may not be the same as that of a capitalization task-specific model. In other words, Single-BiRNN may have different settings selected based on the task at hand. Similarly, for Corr-BiRNN, different settings give best validation SER performance on punctuation and capitalization tasks.

%% file: tables.tex

\begin{table*}[!h]
\begin{adjustbox}{width=1\textwidth}
\centering
\begin{tabular}{|r|c|c|rrr|crr|crr|crrr|}
\hline
                      & \textbf{Task}                            & \textbf{Model}                         & \multicolumn{13}{c|}{\textit{Class Labels}}                                                                                                                                                                                                                                                                                                                                   \\ \hline
\multirow{4}{*}{Ref.} & \multirow{4}{*}{\textit{Punctuation}}    & \multicolumn{1}{r|}{\multirow{2}{*}{}} & \multicolumn{3}{c|}{\textbf{COMMA}}                                         & \multicolumn{3}{c|}{\textbf{PERIOD}}                                                  & \multicolumn{3}{c|}{\textbf{Q-MARK}}                                                   & \multicolumn{4}{c|}{\textbf{OVERALL}}                                                                           \\ \cline{4-16} 
                      &                                          & \multicolumn{1}{r|}{}                  & \multicolumn{1}{c}{P} & \multicolumn{1}{c}{R} & \multicolumn{1}{c|}{F${}_{1}$} & P                               & \multicolumn{1}{c}{R} & \multicolumn{1}{c|}{F${}_{1}$} & P                               & \multicolumn{1}{c}{R} & \multicolumn{1}{c|}{F${}_{1}$} & P                               & \multicolumn{1}{c}{R} & \multicolumn{1}{c}{F${}_{1}$} & \multicolumn{1}{c|}{SER} \\ \cline{3-16} 
                      &                                          & Single-BiRNN                           & 43.7                    & \textbf{54.9}           & \textbf{48.7}           & \multicolumn{1}{r}{\textbf{73.9}} & 19.3                    & 30.6                    & \multicolumn{1}{r}{\textbf{52.3}} & 23.7                    & 32.6                    & \multicolumn{1}{r}{48.0}          & 39.0                    & 43.0                   & 77.6                     \\
                      &                                          & \textbf{Corr-BiRNN}                    & \textbf{57.9}           & 34.3                    & 43.1                    & \multicolumn{1}{r}{62.0}          & \textbf{53.3}           & \textbf{57.3}           & \multicolumn{1}{r}{45.8}          & \textbf{25.7}           & \textbf{32.9}           & \multicolumn{1}{r}{\textbf{59.7}} & \textbf{42.0}           & \textbf{49.3}                   & \textbf{68.9}            \\ \hline
\multirow{4}{*}{Ref.} & \multirow{4}{*}{\textit{Capitalization}} & \multicolumn{1}{r|}{\multirow{2}{*}{}} & \multicolumn{3}{c|}{\textbf{UPPERCASE}}                                     & \multicolumn{3}{c|}{\textbf{SENTENCE-CASE}}                                         & \multicolumn{3}{c|}{\textbf{SINGLE-CASE}}                                             & \multicolumn{4}{c|}{\textbf{OVERALL}}                                                                           \\ \cline{4-16} 
                      &                                          & \multicolumn{1}{r|}{}                  & P                     & \multicolumn{1}{c}{R} & \multicolumn{1}{c|}{F${}_{1}$} & P                               & \multicolumn{1}{c}{R} & \multicolumn{1}{c|}{F${}_{1}$} & P                               & \multicolumn{1}{c}{R} & \multicolumn{1}{c|}{F${}_{1}$} & P                               & \multicolumn{1}{c}{R} & \multicolumn{1}{c}{F${}_{1}$} & \multicolumn{1}{c|}{SER} \\ \cline{3-16} 
                      &                                          & Single-BiRNN                           & \textbf{96.5}                    & \textbf{63.2}           & \textbf{76.4}           & \multicolumn{1}{r}{\textbf{87.0}} & 55.7                    & 67.9                    & \multicolumn{1}{r}{\textbf{99.9}} & \textbf{98.2}           & \textbf{99.0}           & \multicolumn{1}{r}{\textbf{89.6}}          & 61.5                    & 72.9                   & 45.3                     \\
                      &                                          & \textbf{Corr-BiRNN}                    & 95.1                    & \textbf{63.2}           & 76.0                    & \multicolumn{1}{r}{80.9}          & \textbf{65.3}           & \textbf{72.3}           & \multicolumn{1}{r}{99.7}          & 98.0                    & 98.9                    & \multicolumn{1}{r}{84.2}          & \textbf{69.5}           & \textbf{76.2}          & \textbf{43.0}            \\ \hline
\end{tabular}
\end{adjustbox}
\caption{Intelligence Squared (IQ2) results}\label{tab:iq2}
\vspace{-1.5em}
\end{table*}

\begin{table*}[!h]
\centering
\begin{minipage}{\textwidth}   
\begin{adjustbox}{width=1\textwidth}
\label{my-label}
\small
\begin{tabular}{|l|l|lll|lll|lll|llll|}
\hline
\multirow{2}{*}{}                          & \multirow{2}{*}{Model}    & \multicolumn{3}{c|}{\textbf{COMMA}}                       & \multicolumn{3}{c|}{\textbf{PERIOD}}                       & \multicolumn{3}{c|}{\textbf{Q-MARK}}                          & \multicolumn{4}{c|}{\textbf{OVERALL}}                                           \\ \cline{3-15} 
                                           &                           & P           & R                 & F${}_{1}$                  & P                 & R           & F${}_{1}$                  & P                 & R                 & F${}_{1}$                  & P           & R                 & F${}_{1}$                  & SER                 \\ \hline
\multirow{5}{*}{Ref.}                      & Ueffing et al.\cite{ueffing13}            & (45.0)        & (\textbf{47.0})              & (46.0)              & (54.0)              & (72.0)        & (62.0)               & (53.0)              & (33.0)              & (41.0)              & (47.8)         & (54.8)              & (51.0)              & -                   \\
                                           & T-BRNN \cite{tilk2016bidirectional}                  & 64.4          & 45.2                & 53.1                & 72.3                & 71.5          & 71.9                & 67.5                & 58.7                & 62.9                & 68.9          & 58.1                & 63.1                & 51.3                \\
                                           & T-BRNN-pre \cite{tilk2016bidirectional}                & \textbf{65.5} & 47.1                & 54.8                & 73.3                & \textbf{72.5} & 72.9                & \textbf{70.7}       & \textbf{63.0}       & \textbf{66.7}       & \textbf{70.0} & 59.7                & 64.4                & \textbf{49.7}       \\ \cline{2-15}
                                           & Single-BiRNN & \specialcell{62.2\\(\textbf{58.1})}    & \specialcell{47.7\\(41.4)}          & \specialcell{54.0\\(48.4)}          & \specialcell{74.6\\(72.2)}          & \specialcell{72.1\\(72.0)}    & \specialcell{\textbf{73.4}\\(72.1)} & \specialcell{67.5\\(76.9)}          & \specialcell{52.9\\(\textbf{59.5})}          & \specialcell{59.3\\(\textbf{67.1})}          & \specialcell{69.2\\(\textbf{66.1})}    & \specialcell{59.8\\(55.5)}          & \specialcell{64.2\\(60.3)}          & \specialcell{51.1\\(\textbf{58.1})} \\ 
\cline{2-15}
                                                                                   & \textbf{Corr-BiRNN}                & \specialcell{60.9\\(55.6)}    & \specialcell{\textbf{52.4}\\(44.5)} & \specialcell{\textbf{56.4}\\(\textbf{49.4})} & \specialcell{\textbf{75.3}\\(\textbf{72.5})} & \specialcell{70.8\\(\textbf{72.2})}    & \specialcell{73.0\\(\textbf{72.4})}          & \specialcell{\textbf{70.7}\\(74.6)} & \specialcell{56.9\\(56.0)}          & \specialcell{63.0\\(63.9)}          & \specialcell{68.6\\(64.5)}    & \specialcell{\textbf{61.6}\\(\textbf{57.1})} & \specialcell{\textbf{64.9}\\(\textbf{60.6})} & \specialcell{50.8\\(59.2)}          \\    \hline
\multicolumn{1}{|c|}{\multirow{5}{*}{ASR}} & Ueffing et al. \cite{ueffing13}           & -             & -                   & -                   & -                   & -             & -                   & -                   & -                   & -                   & -             & -                   & -                   & -                   \\
\multicolumn{1}{|c|}{}                     & T-BRNN \cite{tilk2016bidirectional}                   & \textbf{60.0} & 45.1                & 51.5                & 69.7                & 69.2          & 69.4                & 61.5                & 45.7                & 52.5                & 65.5          & 57.0                & 60.9                & 57.8                \\
\multicolumn{1}{|c|}{}                     & T-BRNN-pre \cite{tilk2016bidirectional}                & 59.6          & 42.9                & 49.9                & \textbf{70.7}       & \textbf{72.0} & \textbf{71.4}       & 60.7                & 48.6                & 54.0                & \textbf{66.0} & 57.3                & \textbf{61.4}       & \textbf{57.0}                \\ \cline{2-15}
\multicolumn{1}{|c|}{}                     & Single-BiRNN & \specialcell{55.9 \\ (\textbf{45.7})}    & \specialcell{48.7 \\ (35.6)}          & \specialcell{52.0 \\ (40.0)}          & \specialcell{63.1 \\ (60.2)}          & \specialcell{70.9 \\ (\textbf{67.4})}    & \specialcell{66.8 \\ (\textbf{63.6})}          & \specialcell{\textbf{66.7} \\ (\textbf{56.4})}          & \specialcell{\textbf{50.0} \\ (53.7)}          & \specialcell{\textbf{57.1} \\ (55.0)}          & \specialcell{60.1 \\ (\textbf{53.7})}    & \specialcell{59.6 \\ (50.1)}          & \specialcell{59.8 \\ (51.8)}          & \specialcell{64.1 \\ (76.0)}        \\ \cline{2-15} \multicolumn{1}{|c|}{}                   
& \textbf{Corr-BiRNN}                & \specialcell{53.5 \\ (44.9)}    & \specialcell{\textbf{52.5} \\ (\textbf{40.6})} & \specialcell{\textbf{53.0} \\ (\textbf{42.6})} & \specialcell{63.7 \\ (\textbf{61.4})}          & \specialcell{68.7 \\ (64.8)}    & \specialcell{66.2 \\ (63.1)}          & \specialcell{\textbf{66.7} \\ (56.1)}          & \specialcell{\textbf{50.0} \\ (\textbf{56.1})}          & \specialcell{\textbf{57.1} \\ (\textbf{56.1})}          & \specialcell{59.0 \\ (53.2)}    & \specialcell{\textbf{60.3} \\ (\textbf{51.7})} & \specialcell{59.7 \\ (\textbf{52.4})}          & \specialcell{65.4 \\ (\textbf{75.7})}          \\
  \hline
\end{tabular}
\end{adjustbox}
\caption[TED Punctuation Results]{TED punctuation results over test-set-2 and test-set-1 (in parentheses).} \label{tab:ted-punct}
\end{minipage}
\vspace{-1.5em}
\end{table*}

\begin{table*}[!h]
\centering
\begin{adjustbox}{width=1\textwidth}
\label{my-label}
\begin{tabular}{|l|l|lll|lll|lll|llll|}
\hline
\multirow{2}{*}{}     & \multirow{2}{*}{Model} & \multicolumn{3}{c|}{\textbf{UPPERCASE}}        & \multicolumn{3}{c|}{\textbf{SENTENCE-CASE}}    & \multicolumn{3}{c|}{\textbf{SINGLE-CASE}} & \multicolumn{4}{c|}{\textbf{OVERALL}}                      \\ \cline{3-15} 
                      &                        & P         & R        & F${}_{1}$         & P        & R        & F${}_{1}$         & P           & R         & F${}_{1}$         & P        & R        & F${}_{1}$         & SER        \\ \hline
\multirow{2}{*}{Ref.} & Single-BiRNN             & \textbf{94.1} & \textbf{64.0} & \textbf{76.2} & \textbf{84.4} & 68.2 & 75.4 & \textbf{100.0}  & 98.9  & 99.4 & \textbf{88.8} & 75.3 & 81.5 & 33.8 \\
                      & \textbf{Corr-BiRNN}             & 93.7  & 60.0 & 73.2 & 82.6 & \textbf{71.9} & \textbf{76.9} & 99.4   & \textbf{99.7}  & \textbf{99.6} & 87.2 & \textbf{78.2} & \textbf{82.4} & \textbf{33.0} \\
                      \hline
\multirow{2}{*}{ASR}  & Single-BiRNN             & 87.5 & 87.5 & 87.5 & \textbf{80.4} & 58.6 & 67.8 & \textbf{100.0}   & 99.1  & 99.5 & \textbf{86.7} & 69.2 & 76.9 & \textbf{41.3} \\ 
                      & \textbf{Corr-BiRNN}             & \textbf{87.5} & \textbf{87.5} & \textbf{87.5} & 76.3 & \textbf{62.2} & \textbf{68.6} & 99.4    & \textbf{100.0} & \textbf{99.7} & 83.3 & \textbf{72.1} & \textbf{77.3} & 42.3 \\
                      \hline
\end{tabular}
\end{adjustbox}
\caption{TED capitalization results on test-set-2.} \label{tab:ted-case}
\vspace{-1.5em}
\end{table*}

%% file: results.tex
\section{Results and Discussion}

\begin{table*}[!h]
\centering
\label{my-label}
\begin{tabular}{|l|l|l|}
\hline
                             & \textbf{Punctuation}                                                      & \textbf{Capitalization}                                                         \\ \hline
\textit{Gold}                & \begin{tabular}[c]{@{}l@{}}I ended up hiking up Mount Kilimanjaro \textbf{,} the\\ highest mountain in Africa .\end{tabular} & \begin{tabular}[c]{@{}l@{}} I wish you luck . \textbf{May} none of your\\ non cancer cells become endangered species .\end{tabular} \\ \hline
\textit{Single-BiRNN}        & \begin{tabular}[c]{@{}l@{}}i ended up hiking up mount kilimanjaro . the\\ highest mountain in africa .\end{tabular} & \begin{tabular}[c]{@{}l@{}} I wish you luck \textbf{may} none of your\\ non cancer cells become endangered species \end{tabular}     \\ \hline
\textit{\textbf{Corr-BiRNN}} & \begin{tabular}[c]{@{}l@{}}I ended up hiking up Mount Kilimanjaro \textbf{,} the\\ highest mountain in Africa . \end{tabular} & \begin{tabular}[c]{@{}l@{}} I wish you luck . \textbf{May} none of your\\ non cancer cells become endangered species . \end{tabular} \\ \hline
\end{tabular}
\caption{Examples of joint vs. task-specific model predictions on the TED ASR dataset.}\label{tab:anec}
\vspace{-2.0em}
\end{table*}

The test evaluations are reported in Tables 1-3. For all tables, each row contains test evaluation metrics for the hyper-parameter setting that was selected based on validation SER performance of the task being considered. Table~\ref{tab:anec} shows example outputs of our models, on ASR compared to gold labels, created by mapping from reference transcripts. These are shown separately for each of the two tasks for better illustration.

For the IQ2 dataset (see Table~\ref{tab:iq2}), the joint training results in improved performance on both of the tasks, as compared to models trained for each of the individual tasks. This is consistent based on both overall F${}_{1}$-score as well as SER metrics.

For the TED capitalization task (Table~\ref{tab:ted-case}), the Corr-BiRNN model outperforms the Single-BiRNN model performance in terms of F${}_{1}$ score for all test sets across both reference and ASR transcripts (this includes test-set-1 and test-set-2, though results are shown only for test-set-2 in the interest of space). However, improvement is not seen in UPPERCASE performance; this may be explained by the fact that this label does not correlate with any punctuation.

Regarding the TED punctuation task (Table~\ref{tab:ted-punct}), the Corr-BiRNN model outperforms Single-BiRNN (F${}_{1}$ score) for the punctuation task for test-set-1 (Ref.), test-set-1 (ASR) and test-set-2 (Ref.); that is, in three out of four cases. For test-set-2 (ASR), the Single-BiRNN model is marginally better than the Corr-BiRNN model. This is possibly due to the reason that reference transcripts were used for validation (due to unavailability of ASR for validation), because of which the best hyper-parameter setting might have been missed for both models.  

While comparing to the existing baselines on the TED punctuation task (see Table~\ref{tab:ted-punct}), our Corr-BiRNN model fares significantly better on all fronts (especially Q-MARK with 22.9\% gain in F${}_{1}$ score), compared to the existing baseline \cite{ueffing13} for test-set-1 (Ref.). In fact, its performance on test-set-1 (ASR) is better than the baseline for test-set-1 (Ref.). It also outperforms the  T-BRNN \cite{tilk2016bidirectional} baseline in terms of COMMA and PERIOD for test-set-2 (Ref.), which are the more frequent punctuations, in addition to overall, measured in F${}_{1}$ score. For test-set-2 (ASR) though, we do not see improvement, again possibly because the validation dataset is based on reference transcripts.




Despite having a much simpler model, in many cases we were able to beat the baseline performance of T-BRNN \cite{tilk2016bidirectional}, a more complex attention-based BiRNN model. 
This substantiates our claim that joint learning helps learning better representations than task-specific training for a particular task. Our simpler model has the added value of learning and predicting much faster than T-BRNN. In addition, our predictions are generated in one shot over the whole consolidated test sequence and does not need to follow a window based prediction as in T-BRNN.

%% file: related.tex
\section{Related Work}

Simple approaches for single sequence labeling include unigram \cite{lita2003truecasing} and $n$-gram language models \cite{gravano2009restoring}. These models see limited fixed context around a word which may not be sufficient for prediction and they also face data sparsity issues as $n$ increases. There are also classical approaches like Hidden Markov Models (HMM), maximum-entropy models (Max-Ent) and conditional random fields (CRF), all of which try to model a hidden state sequence corresponding to the observed word sequence as in \cite{wang2012dynamic,ma-hovy:2016:P16-1,ueffing13,lu2010better,huang2002maximum,baron2002automatic,liu2006enriching}. However, these models are more difficult to train and construction of hand-crafted features is non-trivial.
Models built using deep neural networks (DNNs) \cite{collobert2011natural,lrec16} usually consist of a context window around the word being considered, which is fed to a multi-layer perceptron that extracts different abstractions of features relevant to the sequence-labeling task. 
More recent approaches \cite{tilk2015lstm,ma-hovy:2016:P16-1,tilk2016bidirectional} considered RNNs, especially LSTMs and reported good results. Compared to the fixed window-based approaches, LSTMs can work on the full sequence of words and dynamically adapt their internal representations. These papers have shown deep learning based solutions outperform the classical approaches.

Multiple sequence labeling tasks and their inter-dependence has been studied in great detail \cite{collobert2011natural}. However, for tasks like POS tagging, NER and chunking, they assumed the availability of punctuation and capitalization, which is not true for ASR transcripts. 
More recently, joint prediction of punctuation and capitalization for transcribed speech has been attempted in \cite{gravano2009restoring}, albeit using $n$-gram language models. In \cite{baldwin2009restoring}, a joint label space for punctuation and capitalization tasks is created, in order to predict labels for both tasks. This is, however, not scalable since label space can possibly explode with the introduction of more labels for each task. A few other works related to joint sequence labeling include joint parsing and punctuation prediction \cite{zhang2013punctuation} using a CRF-based model, and disfluency detection alongside other NLP tasks like punctuation prediction \cite{baron2002automatic} and dependency parsing \cite{honnibal2014joint}, using classical solutions. 
In our work, we explore the joint learning of correlated sequence labeling tasks like punctuation and capitalization using a deep-learning based approach without any feature engineering being involved.

%% file: conclusion.tex
\section{Conclusion}
In this paper, we have shown the utility of models jointly trained on two correlated tasks, punctuation and capitalization, to learn better representations for each of them. Our simple jointly-trained BiRNN model, trained only on lexical features, outperforms several complex models, which demonstrates its robustness and generalization ability. Future work will involve the joint training of a variety of other correlated NLP tasks like POS tagging and NER.